\documentclass{llncs}

\usepackage{algorithm, algorithmic, graphicx, amssymb, amsmath, cite}
\usepackage[english]{babel}

\begin{document}
\title{Density-based Denoising of Point Cloud}
\author{Faisal~Zaman, Ya~Ping~Wong, Boon~Yian~Ng}
\institute{Faculty of Computing and Informatics\\
Multimedia University, Malaysia}

\maketitle

\begin{abstract}

Point cloud source data for surface reconstruction is usually contaminated with noise and outliers. To overcome this deficiency, a density-based point cloud denoising method is presented to remove outliers and noisy points. First, particle-swam optimization technique is employed for automatically approximating optimal bandwidth of multivariate kernel density estimation to ensure the robust performance of density estimation. Then, mean-shift based clustering technique is used to remove outliers through a thresholding scheme. After removing outliers from the point cloud, bilateral mesh filtering is applied to smooth the remaining points. The experimental results show that this approach, comparably, is robust and efficient.
	
\end{abstract}

\section{Introduction}
In the past few decades, surface reconstruction has gained significant attention due to the availability of commodity structured light sensors such as Microsoft Kinect. Point cloud data acquired from these devices is usually contaminated with noise and outliers \cite{berger2014state}, both of which could be caused by the lighting or reflective nature of the surface or artifact in the scene. In recent years, a number of point cloud denoising methods have been proposed  \cite{berger2014state, berger2013benchmark}. Most of the methods work only on certain type of noise or outliers. Denoising algorithms such as data clustering \cite{song2010boundary, sotoodeh2006outlier} are robust to outliers but require prior knowledge about the input objects. Majority voting method \cite{wang2015outlier} can detect several types of outliers cluster, but high computation time makes it infeasible for large datasets. Smoothing algorithms such as moving least square \cite{fleishman2005robust} and mean curvature flow \cite{desbrun2000anisotropic} are also used to remove noise and outliers. These methods treat outliers as point with large noise and project the noisy points to an estimated surface. However, these methods are sensitive against large number of outliers and oversmooth the data points.

In this paper, a point cloud denoising method to remove noise and outliers from the data is proposed. This method consists of the following steps. First, the density of the input data point using kernel density estimation (KDE) technique is evaluated. The performance of KDE is highly influenced by the choice of smoothing parameter, also known as bandwidth. Most of the bandwidth selection methods such as \cite{hyndman2004bandwidth, guidoum2013kernel, zhang2006bayesian, duong2005convergence} involved brute-force or exhaustive search strategies have expensive computation. A particle swarm optimization (PSO) (proposed by Kennedy and Eberhart, 1995) aided bandwidth selection criterion using leave-one-out cross-validation (LOOCV) for multivariate kernel density estimation is proposed. Because of the simple implementation and quick convergence rates of PSO, it has been adapted in many practical applications \cite{blum2015swarm}. The optimal bandwidth selection method ensures the robust performance of density estimation. After estimating the density, the mean-shift clustering algorithm is applied to detect the local maxima of the constructed density estimation. One of the main advantages of using mean-shift algorithm is its ability to localize cluster modes automatically without any prior knowledge of the number of clusters. Once clustering is done, outliers can be removed by the proposed thresholding scheme. After removing the outliers, bilateral mesh filtering is applied to smooth out the rest of the points. Experiments on various point cloud datasets show that the method used in this research gives better result than that of other approaches.

\section{Our Method}
In this section, an overview of the optimal bandwidth selection framework is provided, starting from a review of the classical bandwidth selection methods.

\subsection{Bandwidth in Kernel Density Estimation}

Kernel density estimation (KDE), also known as Parzen window estimation, is a non-parametric way of estimating the underlying distribution or probability density function for a data set. Given $n$ data points $X_{i}$, $i = 1, ..., n$, the kernel density estimate obtained with multivariate kernel function $K_H(x)$ and bandwidth or smoothing parameter $H$, which is a symmetric positive definite matrix, computed at the point $x$ is defined as:

\begin{equation} \label{eq:1}
\hat{f}_{H}(x) = \frac{1}{n}\sum_{i=1}^{n}{K_{H}(x-X_i)}
\end{equation}

where $x=(x_1,x_2,x_3)^T$ is the $3$-dimension vector, $K_H(x) = |H|^{-1/2}K(H^{-1/2}x)	$. The resulting density through KDE depends more heavily on the choice of bandwidth than that of the kernel. For the same data, different bandwidths can produce different results. Optimal bandwidth matrix --- which minimizes the error between the estimated density $\hat{f}_H(x)$ and the true density $f(x)$ is estimated. The performance of  $\hat{f}_H(x)$ can be determined by the risk function or integrated mean squared error (IMSE) such as,

\begin{equation} \label{eq:2}
R(\hat{f}_H(x), f(x)) = E(L(\hat{f}_H(x), f(x)))
\end{equation}

where, $L$ is the loss function and $L(\hat{f}_H(x), f(x)) =\int{[\hat{f}_H(x) - f(x)]^2dx}$. The optimal bandwidth can be obtained by minimizing the risk function, but since the true density $f(x)$ is unknown leave-one-out cross-validation is used to estimate the risk function. Therefore, loss function can be expressed as a function of $H$.

\begin{equation} \label{eq:3}
L(H) = \frac{1}{n} \sum_{i=1}^n log \, \hat{f}_{H,i}(x_i)
\end{equation}

where, $\hat{f}_{H,i}(x_i)$ is the leave-one-out estimator, $\hat{f}_{-i}(x_i)=\frac{1}{(n-1)}\sum_{\substack{j=1 \\ j\neq i}}^{n}{K_H(x_i-x_j)} $ denotes the density estimation of $x_i$ by using the other $(n-1)$ observed data points.  Then the optimal bandwidth matrix $H^*$ can be estimated by,

\begin{equation} \label{eq:4}
H^* = arg \, \underset{H}{max} \, \frac{1}{n} L(H)
\end{equation}

Several classes of parameterizations of the bandwidth matrix are available in the literature. Some popular choices include full bandwidth matrix, a diagonal matrix with positive elements, or using a single bandwidth. In the application, the diagonal bandwidth matrix, $H=diag(h_1^2, h_2^2, h_3^2)$, is employed since it is computationally more efficient than the full bandwidth matrix.

\begin{figure}[t] 
\centering
\includegraphics[width=1\textwidth]{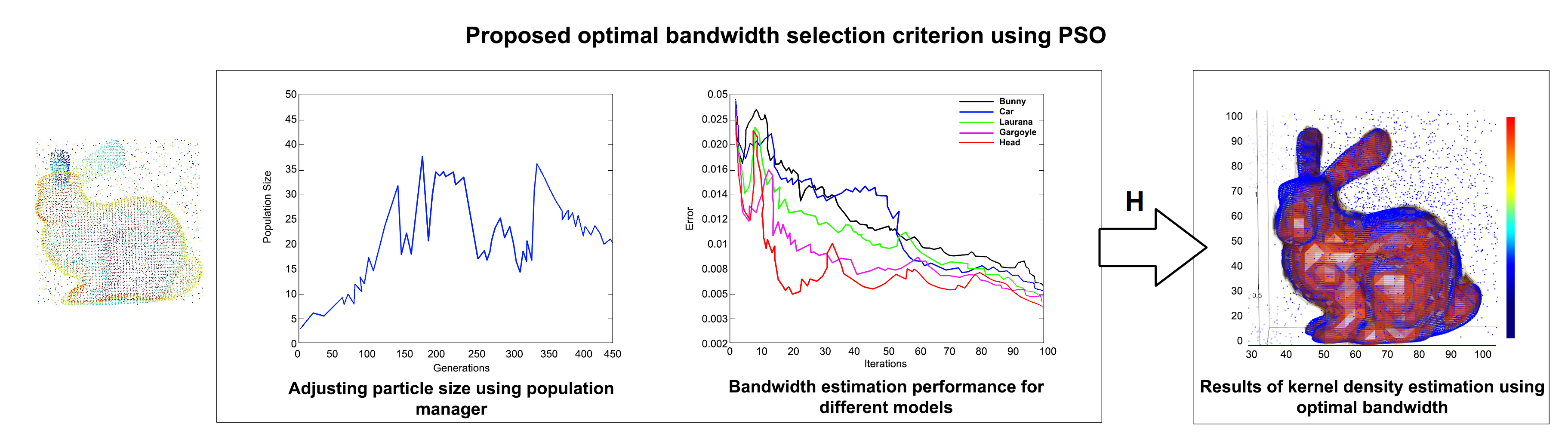}
\caption{The pipeline of the optimal bandwidth selection method, using PSO.}
 \label{fig:pipeline}
\end{figure}

\subsection{PSO for Optimal Bandwidth Selection}
The PSO technique is proposed to carry out the bandwidth selection for better approximation of kernel density estimation. PSO is a nature-inspired meta-heuristic non-linear stochastic optimization technique. PSO is popular for its ability to optimize complex non-linear functions and for its simple implementation. PSO consists of a swarm of particles that fly through hyperspace of the of objective function's landscape. The pipeline of the optimal bandwidth selection method shown in Fig. \ref{fig:pipeline}. The optimization problem is solved from Eq. \ref{eq:4} using PSO, where $H = [h_1, h_2, h_3]^T$ is a $3$-dimensional vector to be optimized and $L(\bullet)$ is the cost function. Here are the steps of PSO algorithm to find the optimal bandwidth; where $H_i^t$ is the potential solution of the $i^{th}$ particle at iteration $t$.

\begin{enumerate}
\item {\textbf{Swarm Initialization:} Set the iteration index $t=0$.  $S$ is the total number of particles, where $\{H_i^t\}_{i=1}^S$} randomly generated particles in solution space.
\item {\textbf{Swarm Evaluation:} Each particle denoted as $p_i^t$ remembers its best position visited so far, which provides the cognitive information. Every particle denoted as $p_g^t$ also knows the best position visited so far among the entire swarm, which provides the social information. The cognitive information $p_i^t$ and the social information $p_g^t$ are updated at each iteration as follows:}

\begin{algorithmic}

 \IF{$L(H_i^t) < L(p_i^{t-1}) $} 
 \STATE{$p_i^t\gets H_i^t$} 
 \ENDIF
 
  \IF{$L(p_i^t) < L(p_g^{t-1}) $} 
 \STATE{$p_g^t\gets p_i^t$} 
 \ENDIF
 
\end{algorithmic}

\item{\textbf{Swarm Update:} Each particle $H_i^t$ has a velocity, denoted as $v_i^t$. The velocity and position of the $i^{th}$ particle are updated in each iteration according to:}

\begin{equation}
	v_i^t = \omega{v_i^{t-1}}+C_cU[0,1](H_i^{t-1}-p_i^{t-1})+C_sU[0,1](H_i^{t-1}-p_g^{t-1})
\end{equation}
\begin{equation}
	H_i^t = H_i^{t-1}+v_i^t
\end{equation}
Here $U[0,1]$ uniform random number between $0$ and $1$ and inertia weight (proposed by Shi and Eberhart \cite{shi1998modified}) sets to $\omega = U[0,1]$. $C_c$ and $C_s$ are constants and called cognitive learning rate and social learning rate, respectively. It controls a particle’s learning behavior. Large value of $C_c$ causes particles scatter around and slow convergence. On the other hand, large value of $C_s$ causes fast convergence, and sometimes it can cause the swarm to converge to local optima. These parameters are chosen carefully so that global optima can be attained without compromising convergence speed. In this application, the parameters $C_c$ reduced from, $2.5$ to $0.5$ and $C_s$ varied between $0.5$ and $2.5$ according to $C_c = 	2.5 - 2t/T$ and $C_s = 0.5 + 2t/T$ \cite{ratnaweera2004self}.

\end{enumerate}

The global best position $p_g$ of all particles during the previous three steps is defined as $  p_g = arg \, \underset{p_i}{max} [L(H_i)]   $, where $p_i = H_i$. Once the maximum number of iterations reached algorithm stopped and return the solution $p_g$.

\begin{figure}[t] 
\centering
\includegraphics[width=1\textwidth]{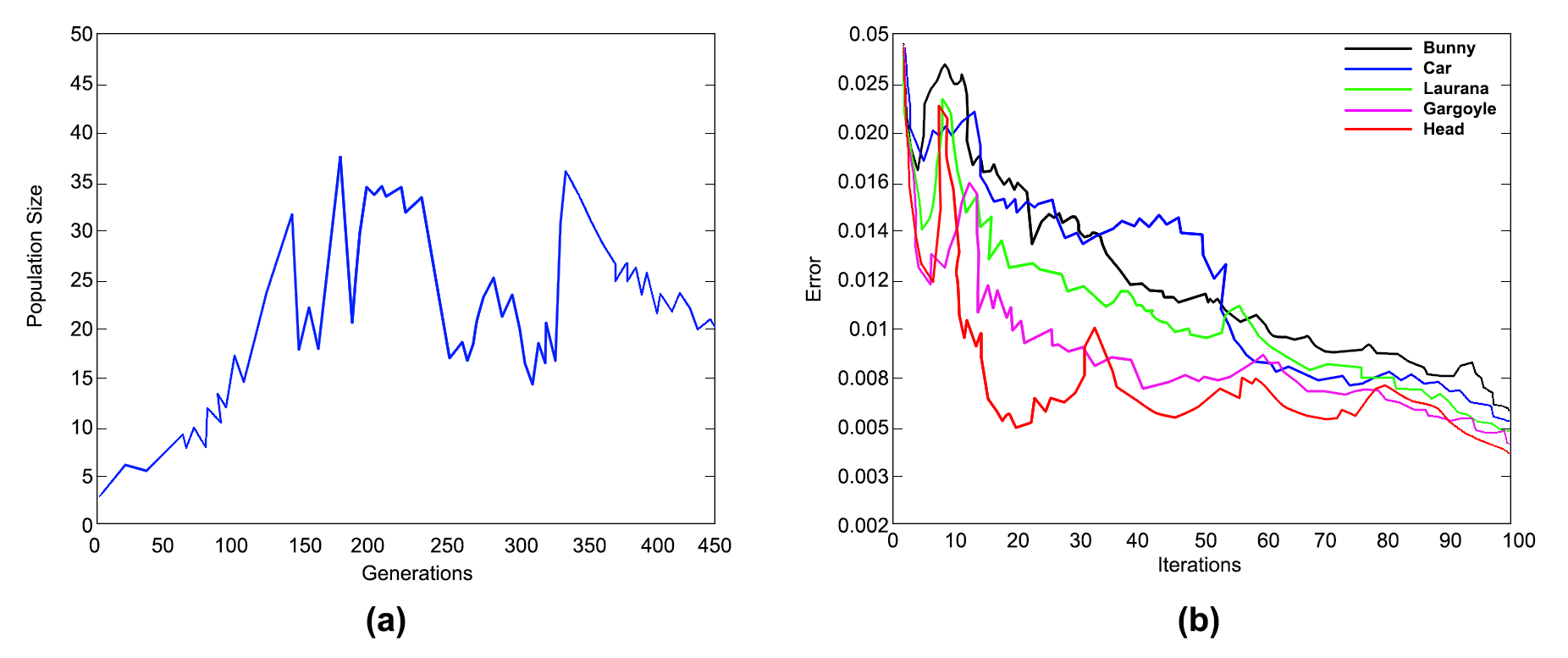}
\caption{(a) Change in the number of particles using Population manager. (b) Estimation performance of our optimal bandwidth selection method.}
 \label{fig:poperror}
\end{figure}

Searching the solution depends highly on the number of selected particles. The number of particles increases the chances of finding global optimal solution, but it also increases the computation expenses significantly. To identify the number of particles suitable for solving the current problem, a method also known as population manager \cite{hsieh2009efficient} is adopted. The idea behind the population manager is to adjust the total number of particles in PSO according to the solution-search status. The population manager determines the number of suitable particles in the following ways:
Assuming $k$ is population-manager activating threshold. For $k$ consecutive generations, if there is no update in $p_g$ and current population size does not exceed the maximum population size, then a new particle will be added to the swarm. This new particle will be added by combining the information from randomly selected two particles in previous best solutions. If population size is equal to the maximum population and there is no update in $p_g$, then remove a particle with poor performance to make space for new potential particle. On the other hand, in $k$ consecutive generations, if $p_g$ update multiple times, then remove a particle with poor performance. Fig. \ref{fig:poperror}a. demonstrates how population manager dynamically adjust the particle in order to facilitate the searching process.

The initial population size in this application is set as one, and the maximum population size is set as 45. To evaluate the PSO based bandwidth selection method, a method to different types of point cloud models is applied (Fig. \ref{fig:poperror}b). As the number of iterations increases, the mean square error (MSE) gradually converges to zero and becomes steady once the optimal bandwidth is found.

\section{Noise and Outliers Removal}
\subsection{Algorithm for Outlier Removal}
Mean-shift-based clustering algorithm \cite{comaniciu2002mean}  is a non-parametric unsupervised clustering technique that does not require any prior information about the number of clusters. It is applied to the kernel density estimate to identify each local maxima or mode  which represents one cluster. The mode of the density function are located at the zeros of the gradient function $\bigtriangledown{f}(x) = 0.$ 
For each data point, say $x$, the mean-shift procedure generates a sequence of points $\{y_{j}\}_{j=1,2,...}$ with
\begin{equation}
y_{j+1} = \frac{\sum_{i=1}^n{x_ig\Big( \Big\| \frac{y_j-x_i}{h_i} \Big\|^2\Big)}}{\sum_{i=1}^n{g\Big( \Big\| \frac{y_j-x_i}{h_i} \Big\|^2\Big)}} \qquad  j = 1,2,....
\end{equation}
and $y_{1} = x $. This sequence converges to a mode of density. The point $x$ belongs to the cluster corresponding to this mode.

\begin{figure}[h] 
\centering
\includegraphics[width=1\textwidth]{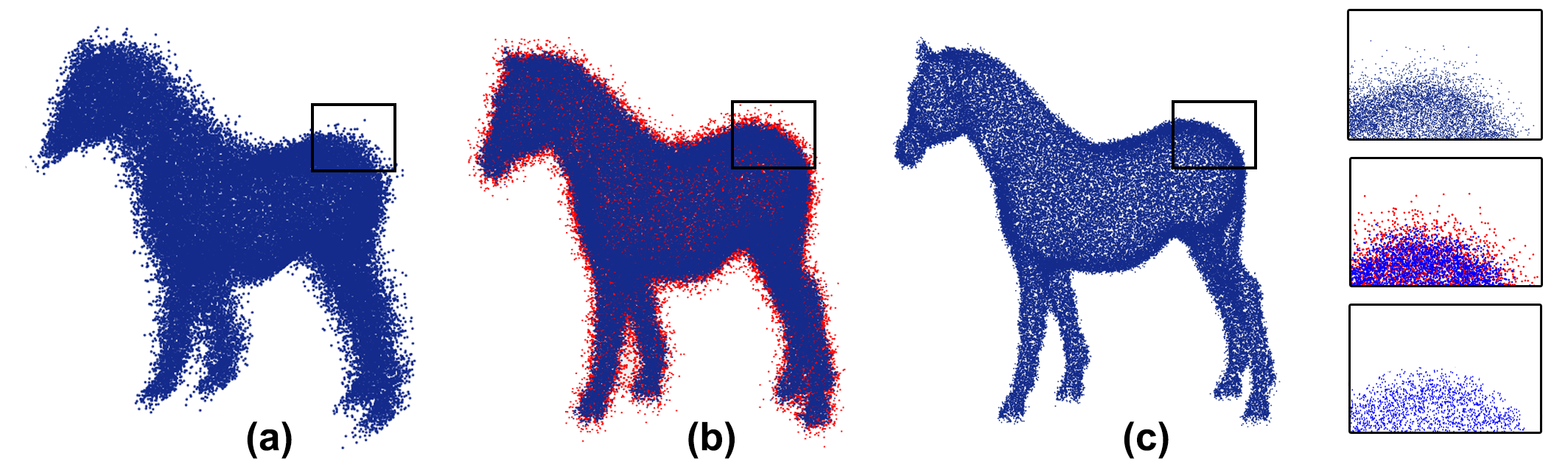}
\caption{Results of outliers removal on horse model. (a) The original point cloud with outliers (b) The outliers detected by our method demonstrated in red color, (c) The resultant point cloud after removing outliers (see the back side of the horse in zoom-views)}
 \label{fig:horse}
\end{figure}

\begin{figure}[h] 
\centering
\includegraphics[width=1\textwidth]{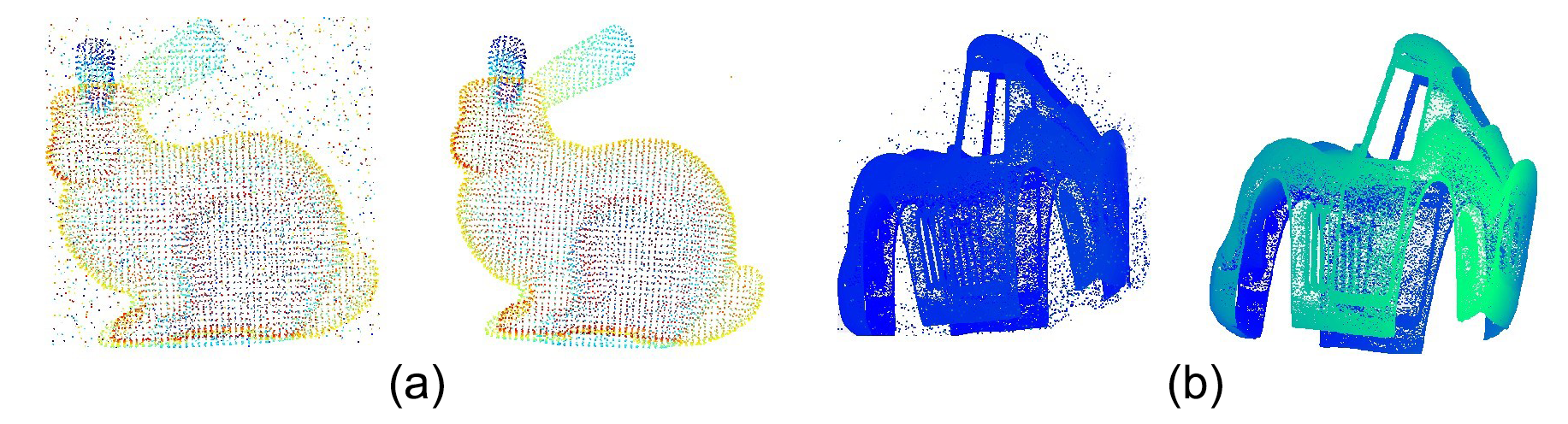}
\caption{Results of the outliers removing algorithm: (a) Left: Bunny point cloud with uniformly distributed noise, Right: Corresponding point cloud after using the method used in this research, (b) Point cloud model of a car with outliers data, Right: Car model after removing outliers.}
 \label{fig:outliers}
\end{figure}

After the clustering is done, the outliers need to be determined. For each point $p$ in a cluster, where $x_p$ is the original location of $p$, the average distance ($\bar{x}$) of all its neighboring points using k-nearest neighbor(kNN)is calculated. Then the point $p$ is shifted iteratively to a new position $\bar{x}$. After that, the distance between two positions, $||\bar{x}$ - $x_p||$, is calculated. If the difference between $||\bar{x}$ - $x_p||$ and the average distance between its shifted neighbors below a certain threshold, then point $p$ is considered outlier (Fig. \ref{fig:horse}b). The outliers are discarded, leaving behind a new set of points, which have better representation of the true shape of the cluster. Fig. \ref{fig:outliers} demonstrates the results of the outliers removal algorithm in different models.

\subsection{Experimental Results and Analysis}
In order to validate the performance and effectiveness of the proposed method, the estimated performance in both synthetic and real world dataset is tested. This program is written in Python, and the data obtained from the research is from the tests running on Intel Core-i7 CPU 860 at 2.80 GHz plus 4GB RAM PC. Fig. \ref{fig:outliers} demonstrates some of the results of this method(standford bunny and car models). To evaluate the applicability of this method, both raw and synthetic noise was tested. Table \ref{table:computation} summarizes timings and the parameters used to generate the results. $T_h$ and $T_f$  is the computation time for the optimal bandwidth selection using PSO and outliers removal using thresholding scheme respectively. 

\begin{table}[!h]
\caption{Computation time of the optimal bandwidth selection and outliers removal method for various models.}
\label{table:computation}
\begin{center}
\renewcommand{\arraystretch}{1.4}
\setlength\tabcolsep{3pt}
\begin{tabular}{llllll}
\hline\noalign{\smallskip}
 Data & Input points & Filtered points & $T_h$ & $T_f$ \\
\noalign{\smallskip}
\hline
\noalign{\smallskip}
Bunny & 330K & 293K & 29s & 30s \\
Car & 720K & 600K & 1m 45s & 1m 25s  \\
Horse & 364K & 214K & 44.43s & 18.45s  \\
Gargoyle & 2.1M & 796K & 3m 2s & 4m 44s  \\
Head & 1.9M & 1.2M & 2m 5s & 3m 25s  \\
\hline
\end{tabular}
\end{center}
\end{table}
Results of the denoising approach on scanned data show that it has good performance on different types of scanned data. Experiments illustrate the capability of this method in removing outliers. 

\subsection{Algorithm for Noise Removal}
\begin{figure}[!h] 
\centering
\includegraphics[width=1\textwidth]{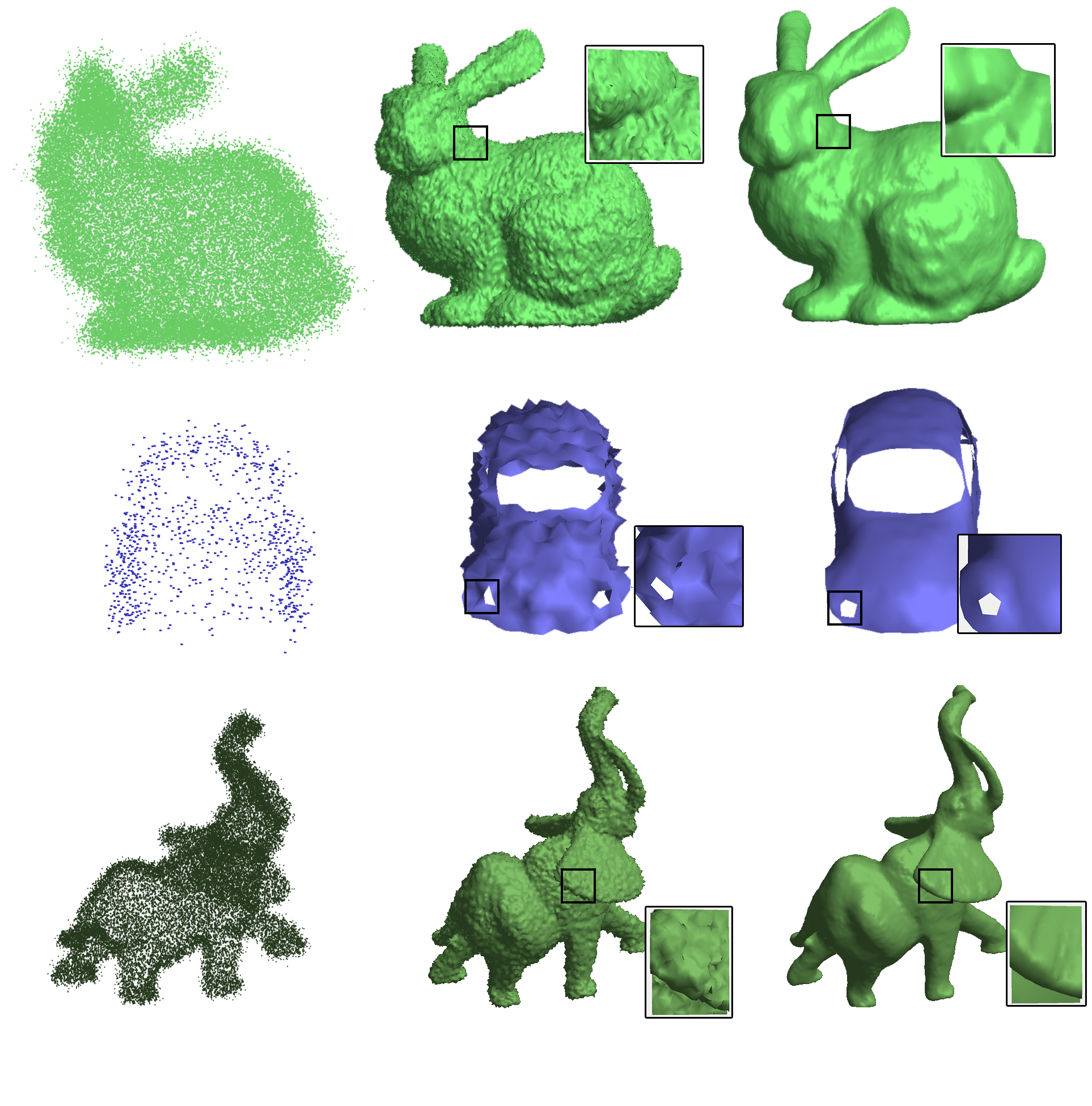}
\caption{Point cloud denosing results: The first column is the point cloud models after removing outliers; second column represents the triangular mesh representation of the point cloud and third column is the models after applying bilateral mesh denoising.}
 \label{fig:denoising}
\end{figure}

After removing outliers, triangular mesh for the rest of the points is calculated. After that, the noise from the mesh by applying bilateral mesh filter \cite{fleishman2003bilateral} is removed. Bilateral mesh filter is popular for its ability to remove noise and preserve features at the same time. For denoising a point $p$, other points around its neighborhood $\{q_i\}$ is identified; where, $||p - q_i|| < \rho = |2\sigma_{c}|$, here $\sigma_c$ is the radius of the neighborhood of point $p$. To determine a desire value for $\sigma_c$, the average distance of all adjacent triangles in an input mesh is used. The bilateral filters in this method also have two parameters, $\sigma_c$ and $\sigma_s$. The $\sigma_s$ is the standard deviation of the offsets in the selected neighborhood. The result of bilateral mesh filtering operation on the input model is illustrated in Fig. \ref{fig:denoising}.

\section{Conclusion}
In this paper, a density-based approach for denoising point cloud data is proposed. In particular, particle swarm optimization technique is used for estimating the optimal bandwidth matrix of kernel density estimation, and mean-shift clustering technique to remove outliers. To remove noise from remaining points, bilateral mesh denoising method is applied. Our method is very robust with highly noisy dataset, as can be seen from the examples shown. While this research illustrates only a few applications of the method within the domain of point cloud data, it is believed that its simplicity and effectiveness will lead to its application in other domain.

\newpage

\bibliographystyle{splncs03}
\bibliography{bibtex}

\end{document}